\documentclass[journal]{elsarticle}

\usepackage[margin=1in]{geometry}
\usepackage{lipsum}
\usepackage{mathtools}

\usepackage{blindtext}

\usepackage[11pt]{moresize}
\usepackage{subfig}
\usepackage{amsmath}
\usepackage{graphicx}
\usepackage{times}
\usepackage{graphicx}
\usepackage{multirow}
\usepackage{breqn}
\usepackage{tabularx}
\usepackage{epstopdf}

\newcolumntype{Y}{>{\centering\arraybackslash}X}

\usepackage{epstopdf}
  \usepackage[norule]{footmisc}

\usepackage{amssymb}
 \usepackage{algorithm, algorithmic}
\usepackage{enumitem}
\usepackage{booktabs,siunitx}
\newcommand\blfootnote[1]{%
  \begingroup
  \renewcommand\thefootnote{}\footnote{#1}%
  \addtocounter{footnote}{-1}%
  \endgroup
}
\usepackage{array}
\usepackage{ifpdf}
\usepackage{mathtools}
\usepackage{lipsum}
\usepackage{amsmath, amsfonts, pifont, float, color, url}
\usepackage{graphicx}
\usepackage[draft]{hyperref}
\newcolumntype{x}[1]{%
>{\centering\hspace{0pt}}m{#1}}%

\usepackage{array}

\usepackage{multirow}

\begin{document}

 \newdimen\origiwspc%
  \newdimen\origiwstr%
  \origiwspc=\fontdimen2\font
  \origiwstr=\fontdimen3\font

\thispagestyle{empty}	
%
\font\myfont=cmr12 at 16pt
\title{{\myfont Role of Awareness and Universal Context in a Spiking Conscious Neural Network (SCNN): A New Perspective and Future Directions}}
\author{Ahsan Adeel}


%
\address{Computing Science and Mathematics, University of Stirling, FK9 4LA, UK    \\
deepCI.org, 20/1 Parkside Terrace, Edinburgh, EH16 5XW, UK
}



\blfootnote{*ahsan.adeel@deepci.org \textcircled{c} Ahsan Adeel 2018. All rights reserved.} 
\begin{abstract}
Awareness plays a major role in human cognition and adaptive behaviour, though mechanisms involved remain unknown. Awareness is not an objectively established fact, therefore, despite extensive research, scientists have not been able to fully interpret its contribution in multisensory integration and precise neural firing, hence, questions remain: (1) How the biological neuron integrates the incoming multisensory signals with respect to different situations? (2) How are the roles of incoming multisensory signals defined (selective amplification or attenuation) that help neuron(s) to originate a precise neural firing complying with the anticipated behavioural-constraint of the environment? (3) How are the external environment and anticipated behaviour integrated? Recently, scientists have exploited deep learning architectures to integrate multimodal cues and capture context-dependent meanings. Yet, these methods suffer from imprecise behavioural representation and a limited  understanding of neural circuitry or underlying information processing mechanisms with respect to the outside world. In this research, we introduce a new theory on the role of awareness and universal context that can help answering the aforementioned crucial neuroscience questions. Specifically, we propose a class of spiking conscious neuron in which the output depends on three functionally distinctive integrated input variables: receptive field (RF), local contextual field (LCF), and universal contextual field (UCF) - a newly proposed dimension. The RF defines the incoming ambiguous sensory signal, LCF defines the modulatory sensory signal coming from other parts of the brain, and UCF defines the awareness. It is believed that the conscious neuron inherently contains enough knowledge about the situation in which the problem is to be solved based on past learning and reasoning and it defines the precise role of incoming multisensory signals (amplification or attenuation) to originate a precise neural firing (exhibiting switch-like behaviour). It is shown, when implemented within an SCNN, the conscious neuron helps modelling a more precise human behaviour e.g., when exploited to model human audio-visual speech processing, the SCNN performed comparably to deep long-short-term memory (LSTM) network. We believe that the proposed theory could be applied to address a range of real-world problems including elusive neural disruptions, explainable artificial intelligence, human-like computing, low-power neuromorphic chips etc.

\end{abstract}


\maketitle
\section{Introduction}
What is awareness or consciousness? Think of a well-trained and experienced car driver who automatically identifies and follows the traffic protocols in different surrounding environments (e.g. street, highway, and city centre) by interpreting the visual scenes directly (such as buildings, school etc.). Similarly, imagine a car with defective parking sensors which sometimes miscalculate the nearby object distance. It means that the audio input is ambiguous and driver can't fully rely on parking sensors for precise maneuvering decisions e.g. while reversing the car. In this situation, we observe that the driver automatically starts utilizing visual cues to leverage the complementary strengths of both ambiguous sound (defective reversing beeps) and visuals. This is one example of consciousness or awareness, where the surrounding environment or situation (UCF) helps establishing the anticipated behaviour to comply with, defining the optimal roles of incoming multisensory information, and eventually controlling human actions. 

Similarly, in contextual audio-visual (AV) speech processing, we observe that in a very noisy environment, our brain naturally utilizes other modalities (such as lips, body language, facial expressions) to perceive speech or the conveyed message (i.e. speech perception in noise) \cite{sumby1954visual}\cite{mcgurk1976hearing}\cite{summerfield1979use}\cite{patterson2003two}. However, it raises crucial questions: How does it happen in the brain? How the incoming sensory signals (such as vision and sound) integrate with respect to the situation? How are the roles of incoming signals (selective amplification/suppression) defined? How does the neuron(s) originate a precise control command that controls human actions based on incoming multisensory information and their precise integration, complying with the anticipated behavioural-constraint of the environment? Certainly, defining the context and its relevant features knowing when a change in context has taken place are challenging problems in modelling human behaviour \cite{gonzalez2008formalizing}. It is also claimed in the literature that context could be of infinite dimensions but humans have a unique capability of correlating the significant context and set its boundaries intuitively with respect to the situation. However, once the context is identified, it is relatively easy to utilize and set its bounds to more precisely define the search space for the selection of best possible decision \cite{gonzalez2008formalizing}. 

A simple example of contextual modulation is shown in Figure 1 \cite{kay2018contrasting}. It illustrates the role of localized contextual information (i.e. LCF) that comes from the nearby location in space. It can be seen that the ambitious RF input (in the top row) is interpreted as B or 13 depending on the local contextual information coming from the nearby location in space. Identically, consider the perception of any ambiguous letter or speech sound. At times, if available, the surrounding environment and its understanding significantly help to disambiguate the ambiguous input. The contextual modulation can in principle be from anywhere in space/time that modulates the transmission of information about other driving signals  \cite{kay2018contrasting}. However, the selective amplification and attenuation of incoming multisensory information with respect to the outside world at the neural level is still very little understood. In addition, to the best of our knowledge, not much progress has been made to fully interpret the role of consciousness and objectively define its contribution in multisensory integration. The complexity of the problem is widely appreciated by scientists with a consensus that it is not easy to use awareness and contextual modulation to show enhanced processing, learning, and reasoning. 

In this research article, we propose a novel conscious neural structure and objectively define awareness in terms of newly proposed UCF. The proposed spiking conscious neuron exhibits a switch-like behavior that defines the role of incoming multisensory signals with respect to the outside environment and anticipated behaviour. It is believed that the conscious neuron inherently contains enough knowledge about the situation in which the problem is to be solved based on past learning and reasoning and it helps defining the precise role of incoming multimodal signals to originate a precise control command. The conscious neuron exploits four types of contexts: modulatory (LCF), temporal, spatial, and awareness (UCF). The preliminary behavioural modelling analysis and simulation results demonstrate enhanced learning and reasoning capability of the proposed SCNN as compared to the state-of-the-art unimodal and multimodal models. 

The rest of the paper is organized as follows: Section 2 discusses the conceptual foundation and motivation which leads to the development of conscious neural structure and SCNN. Section 3 presents the conscious neural structure and SCNN. In section 4, the conscious neural structure and SCNN are utilized for behavioural modelling including AV speech processing and driving behaviour. Finally, section 5 discusses the research impact, applications, and future research directions. 


\begin{figure} 
	\centering
	\includegraphics[trim=0cm 0cm 0cm 0cm, clip=true, width=0.16\textwidth]{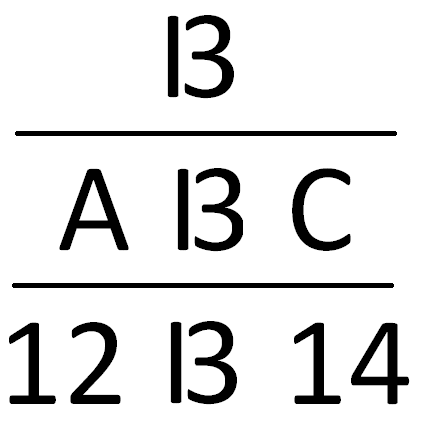}
	\caption{Ambiguous decision-making and local contextual modulation}
	\label{fig:Picture2}
\end{figure}
\section{Motivation and Contribution}
The simplified state-of-the-art integrate-and-fire neural structure is doing wonders today, think of the potential of neuron representing a closer form of biophysical reality. There exists ample evidence that divisive and multiplicative gain modulations are widely spread in mammalian neocortex with an indication of amplification or attenuation via contextual modulation \cite{kay2018contrasting}. Evidence gathered in the literature suggests that multisensory interactions emerge at the primary cortical level \cite{stein2008multisensory}\cite{stein2009neural}. Scientists have presented several theories and empirical results on the role of contextual modulation to disambiguate the ambiguous input or improve feature detection with weak or noisy inputs \cite{kay1998contextually}. Recently, the authors in \cite{kay2018contrasting} used modern developments in the foundation of information theory to study the properties of local processors (neuron or microcircuit) embedded within the neural system that uses the contextual input to amplify or attenuate transmission of information about their driving inputs.  Specifically, the authors used advances in information decomposition to show that the information transmitted by the local processor with two distinct inputs (driving and contextual information) can be decomposed into components unique to each other having three-way mutual/shared information. In \cite{kay1998contextually}, the authors used an edge detection problem as a benchmark to demonstrate the effectiveness of contextual modulation in recognizing specific patterns with noisy RF input. It was shown how surrounding regions in different parallel streams helped detecting the edge within any particular region and played a significant role in combating noisy input. 

Recently, researchers have also proposed several deep recurrent neural network architectures to exploit contextual modulation by leveraging the complementary strengths of multimodal cues. For example, researchers in \cite{ephrat2018looking} presented a joint AV model for isolating a single speech signal from a mixture of sounds. The authors used a complementary strength of both audio and visual cues using deep learning to focus the audio on the desired speaker in a scene. Similarly, the authors in \cite{adeel2018contextual} and \cite{gogate2018dnn} developed an AV switching component for speech enhancement and mask estimation to effectively account for different noisy conditions contextually. The contextual AV switching components were developed by integrating a convolutional neural network (CNN) and long-short-term memory (LSTM) network. However, these end-to-end multimodal learning models operate at the network level and can't be used for deep analysis and information decomposition to understand neural circuitry and underlying information processing mechanisms at the neural level, with respect to the outside world and anticipated behaviour. In addition, these methods only exploit the localized context without considering the overall knowledge of the problem (awareness). Thus, the limited contextual exploitation leads to imprecise behavioural representation. This work proposes a new conscious neural structure and SCNN that objectively define awareness at the neural level. Contrary to the work presented in \cite{kay1998contextually}, the proposed SCNN is evaluated with a noisy speech filtering problem. Specifically, we used two distinctive multimodal multistreams (lip movements as LCF and noisy speech as RF) and studied how the LCF helped to improve the noisy speech filtering in different noisy conditions (ranging from a very noisy to an almost zero noise environment). Later, going beyond the theory of local contextual modulation, we added UCF as another input variable (as a fourth virtual dimension) to define descriptive and control context (environment and anticipated behaviour) in SCNN.

\section{Spiking Conscious Neural Network}
The proposed conscious neural structure is presented in Figure 2. The output of the neuron depends on three functionally distinctive integrated input variables: driving (RF), modulatory (LCF), and awareness (virtual UCF). The RF is defining the ambiguous sensory signal, LCF is defining the modulatory sensory signal coming from other parts of the brain, and UCF is defining the outside world and anticiapted behaviour. The interaction among RF, LCF, and UCF in an SCNN is shown in Figure 3. The output is denoted by the random variable Y, whereas X, Z, and U represent RF, LCF, and UCF respectively. It is believed that the proposed neural structure when implemented within a multi-layered multiunit network of similar neurons produce a widely distributed activity pattern with respect to the current circumstances (i.e. a combination of RF, LCF, and UCF at the neuronal level). This activity helps neural network to explore and exploit the associative relations between the features extracted within different streams \cite{kay1998contextually}\cite{kay2018contrasting}. In the implementation, the neuron in one stream is connected to all other neurons in the neighbouring stream of the same layer. This is achieved through shared connections among the neurons that guide learning and processing with respect to local and universal contexts. 
\begin{figure} [htb!]
	\centering
	\includegraphics[trim=0cm 0cm 0cm 0cm, clip=true, width=0.80\textwidth]{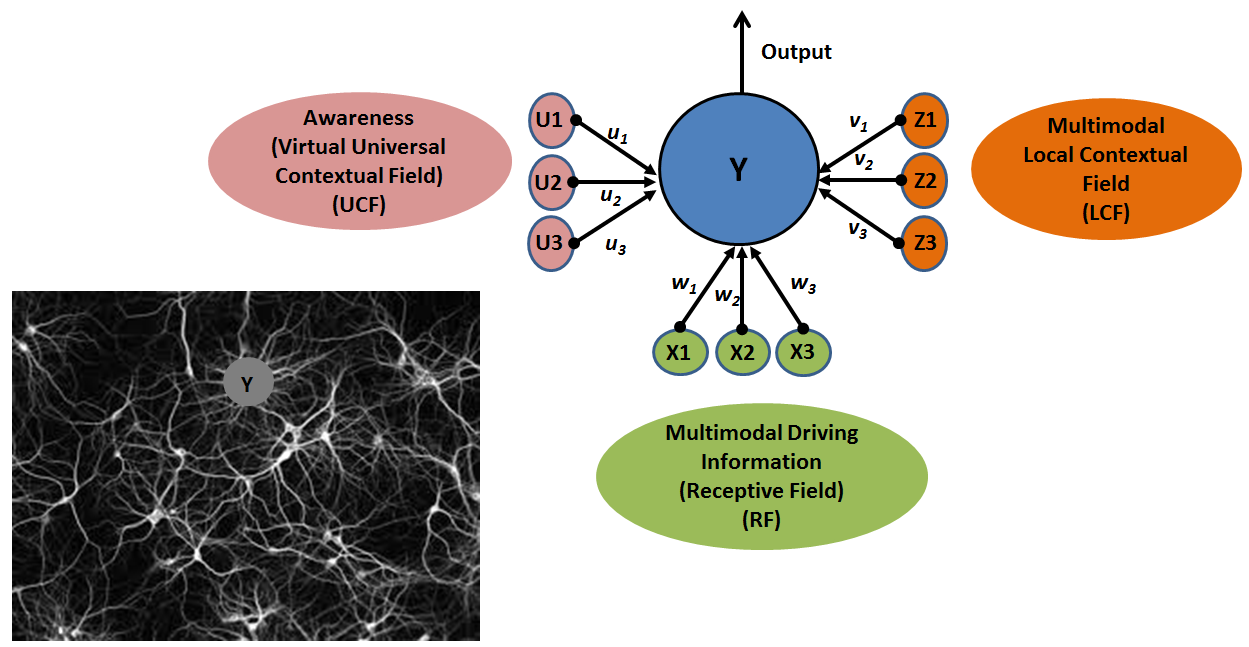}
	\caption{Proposed conscious neural structure with a switch-like behaviour: The output depends on three functionally distinctive integrated input variables: driving (RF), modulatory (LCF), and awareness (virtual UCF). The RF is defining the ambiguous sensory signal, LCF is defining the modulatory sensory signal coming from other parts of the brain, and UCF is defining the outside world and anticiapted behaviour. The conscious neuron, with respect to the outside environment (UCF), decides whether the role of LCF is modulatory or null.}
	\label{fig:Picture2}
\end{figure} 
\subsection{Mathematical Modelling}
The conscious neuron (\textit{y}) in the proposed SCNN interacts by exchanging the excitatory and inhibitory spikes probabilistically (in the form of bipolar signal trains) as shown in Figure 3 and Figure 4. In steady state, the stochastic spiking behaviour of the network has a “product form” property (product of firing rates and transition probabilities) which defines state probability distribution with easily solvable non-linear network equations. The firing from neuron \textit{y} to succeeding neuron \textit{w} in the network is according to the Poisson process, represented by the synaptic weights $w_{yw}^+$ = $r_y[P_{yx}^+ + P_{yz}^+ + P_{yu}^{+}]$ and $w_{yw}^-$ = $r_y[P_{yx}^- + P_{yz}^- + P_{yu}^-]$, where $P_{yx}^+, P_{yz}^+, P_{yu}^+$ and $P_{yx}^-, P_{yz}^-, P_{yu}^-$ represent the probabilities of excitatory and inhibitory RF, LCF, and UCF signals, respectively. The term $r_y$ represents the firing rate of the conscious neuron. The terms $w_{yx}^+$, $w_{yz}^+$, $w_{yu}^+$ and $w_{yx}^-$, $w_{yz}^-$, $w_{yu}^-$ represent the RF, LCF, and UCF synaptic weights (i.e. the rates of positive and negative signal transmission) that network learns through the process of learning or training. In the network, the conscious neuron \textit{y} can receive exogenous signals positive/negative from the inside (within the network) or outside world according to Poisson arrival streams of rates $\Lambda_x$, $\lambda_x$, respectively. The potential (\textit{Y}) of the conscious neuron represents its state that increases/decreases with respect to an incoming signal coming from the inside or outside world. The proposed neural structure is implemented using G-networks which possess product-form asymptotic solution \cite{gelenbe1993g}. The neuron \textit{y} in firing state transmits an impulse to neuron \textit{w} with a Poisson rate ($r_y$) and probability $P^+ (y, w)$ or $P^- (y, w)$ depending on the incoming signal as excitatory or inhibitory. The transmitted signal can also leave the network and go outside the world with probability $d(y)$ such that:

\begin{figure} 
	\centering
	\includegraphics[trim=0cm 0cm 0cm 0cm, clip=true, width=0.75\textwidth]{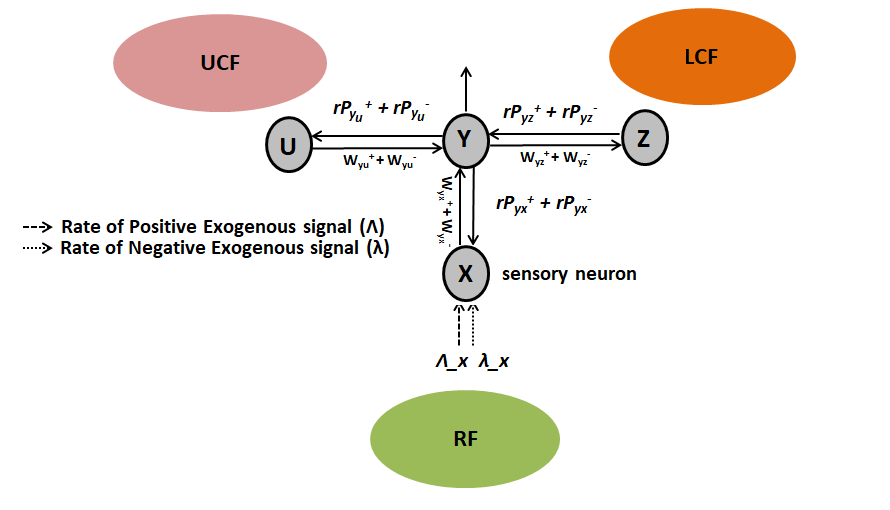}
	\caption{The interaction among RF, LCF, and UCF in an SCNN. Please note that the switch-like behaviour or filtering rules are enforced by the positive and negative synaptic weights associated with each input field.}
	\label{fig:Picture2}
\end{figure} 
\begin{figure} [!htb]
	\centering
	\includegraphics[trim=0cm 0cm 0cm 0cm, clip=true, width=0.65\textwidth]{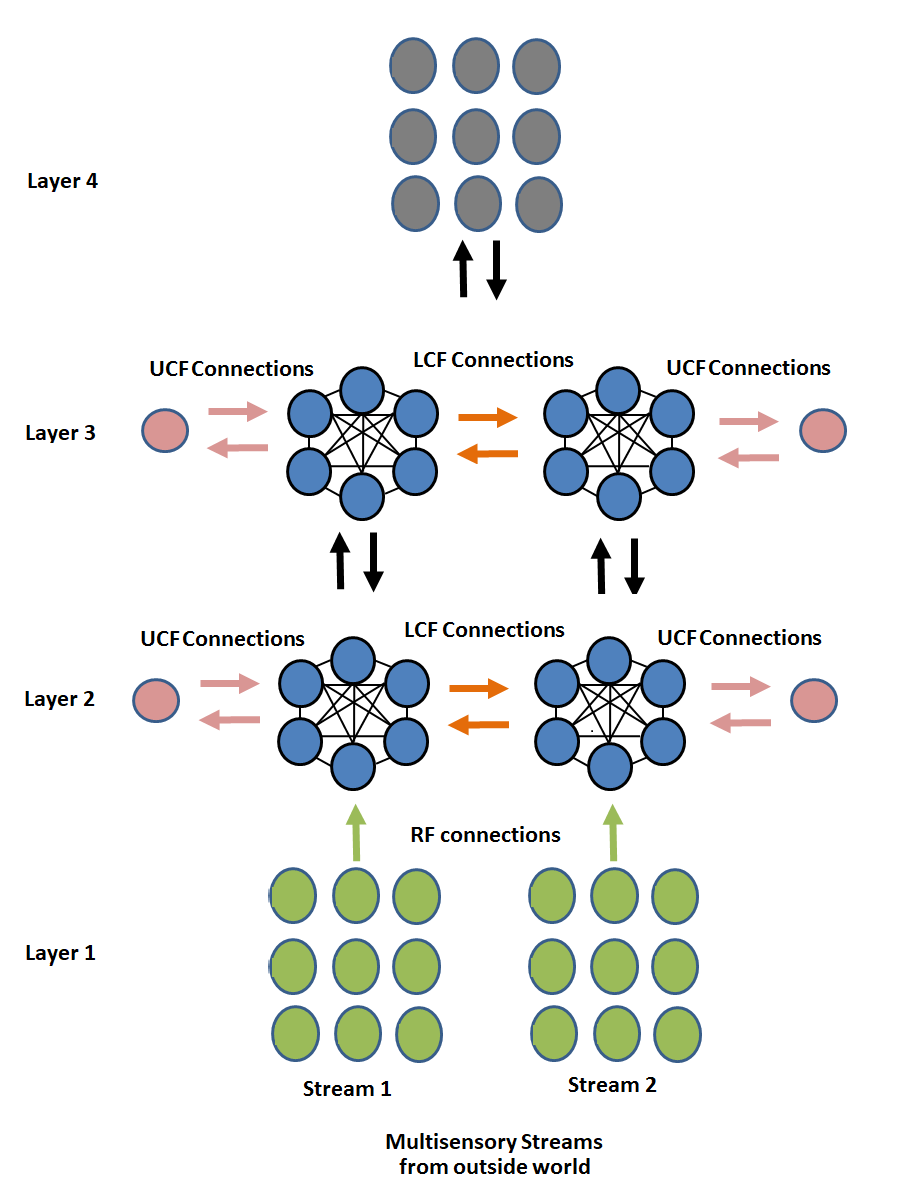}
	\caption{Proposed SCNN: Multi-layered multiunit network of many similar conscious neurons, where the unit in one stream is connected to all other units in the neighbouring streams of the same layer.}
	\label{fig:Picture2}
\end{figure} 

\begin{multline}
d(y) + \sum_{x=1}^{N} [P^+ (y, x) + P^- (y, x)] + \sum_{z=1}^{N} [P^+ (y, z) + P^- (y, z)] + \sum_{u=1}^{N} [P^+ (y, u) + P^- (y, u)] = 1 
\end{multline}

Where, 

\begin{multline}
w^+ (y,w) = r_y[P^+ (y,x) + P^+ (y,z)+P^+ (y,u)]\geq 0, 
w^- (y,w) = r_y[P^- (y,x) + P^- (y,z)+P^- (y,u)]\geq 0  
\end{multline}

The firing rate of the conscious neuron can be written as:
 \begin{multline}
 r(y) = (1-d(y))^{-1} (\sum_{x=1}^{N} [w^+ (y,x) + 
 w^- (y,x)] + \sum_{z=1}^{N} [w^+ (y,z) + w^- (y,z)]  
 + \sum_{u=1}^{N} [w^+ (y,u) + w^- (y,u)])
 \end{multline}
 
If $Y(t)$ is the potential of the conscious neuron \textit{y} then in \textit{n} number of neurons, vector $\overline{Y(t)}$ = $(Y_1(t), Y_2(t), ...,Y_n(t))$ can be modelled as a continuous-time Markov process. The stationary joint probability of the network is given as:

\begin{equation}
\lim\limits_{n\to\infty} P(\overline{Y(t)}) = y_1(t), y_2(t), ...,y_n(t) = \prod_{y=1}^{n}(1-q_y)q_y^{ny}, q_y = \frac{Q_Y^+}{r_y + Q_Y^-}
\end{equation}

Where $Q_Y^+$ and $Q_Y^-$ are the average rate of +ive and -ive signals at neuron \emph y, given as:

\begin{equation}
Q_Y^+ =  \sum_{x=1}^{N}q_x w^+(y, x) + \sum_{z=1}^{N}q_z w^+(y, z) + \sum_{u=1}^{N}q_u w^+(y, u)    
\end{equation}

\begin{equation}
Q_Y^- =  \sum_{x=1}^{N}q_x w^-(y, x) + \sum_{z=1}^{N}q_z w^-(y, z) + \sum_{u=1}^{N}q_u w^-(y, u)    
\end{equation}

The probability that the conscious neuron (\textit{Y}) is excited can be written as:

\begin{equation}
q_y = \frac{\sum_{x=1}^{N}q_x w^+ (y,x) + \sum_{z=1}^{N}q_z w^+ (y,z) + \sum_{u=1}^{N}q_u w^+ (y,u)}{[W_W^+ + W_W^-] + \sum_{w=1}^{N}q_x w^- (y,x) + \sum_{z=1}^{N}q_z w^- (y,z) + \sum_{u=1}^{N}q_u w^- (y,u)} 
\end{equation}

%
%
%
%
%

where $w^+ (y,x)$, $w^- (y,x) $, $w^+ (y,z)$, $w^- (y,z) $ $w^+ (y,u)$, $w^+ (y,u)$ are the positive and negative RF, LCF, and UCF weights. $W^+ _W$ and $W^- _W$ are the positive and negative weights between the conscious neuron \textit{y} and succeeded neuron \textit{w}. For training and weights update, state-of-the-art gradient descent algorithm is used. The RF input ($q_x$) is given as:     

\begin{equation}
q_x = \frac{Q_x^+}{[w(x,y)^+ + w(x,y)^-] + Q_x^-} 
\end{equation}

\begin{equation}
Q_x^+ = \Lambda_x + \sum_{v=1}^{N}q_v w^+(x, v)   
\end{equation}

\begin{equation}
Q_x^+ = \lambda_x + \sum_{v=1}^{N}q_v w^- (x, v)   
\end{equation}

Where $q_v$ is the potential of the preceding neuron \textit{v} coming from the outside world, and $q_u$ and $q_z$ in (7) are the potentials of incoming LCF and UCF. It is to be noted that $w(x,y)^+$ and $w(y,x)^+$ are different.

\subsection{Information Decomposition}
A Venn diagram of the information theoretic measures for distinctive integrated input variables is depicted in Figure 5, where RF, LCF, and UCF are represented by the green, orange, and grayish pink ellipses respectively. The output (Y) is represent by the blue ellipse. In information processing equations, the output is denoted by the random variable Y, whereas RF, LCF, and UCF are represented by X, Z, and U respectively. \\
\begin{figure} [htb]
	\centering
	\includegraphics[trim=0cm 0cm 0cm 0cm, clip=true, width=0.25\textwidth]{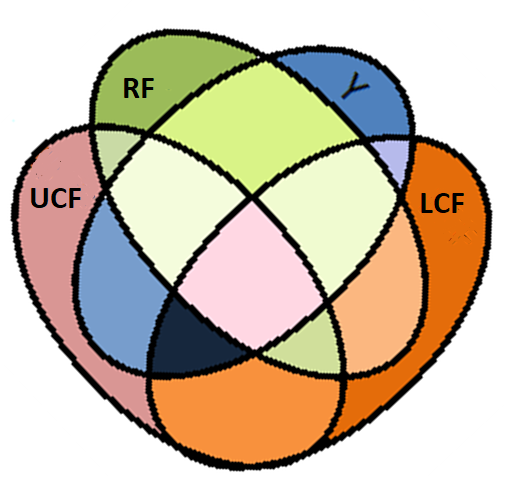}
	\caption{Venn diagram of information theoretic measures for distinctive integrated input variables RF, LCF, and UCF represented by the green ellipse, orange ellipse, and grayish pink ellipse respectively. The output (Y) is represent by the blue ellipse. The UCF (U) and associated $H(U|X,Y,Z)$ is interpreted as the information contained in U but not in X, Y, and Z. The output (Y) and associated $H(Y|X,Z,U)$ is interpreted as the information contained in Y but not in X, Z, and U. The LCF (Z) and associated $H(Z|X,Y,U)$ is interpreted as the information contained in Z but not in X, Y, and U. The RF (X) and associated $H(X|Y,Z,U)$ is interpreted as the information contained in X but not in Y, Z, and U.} 
	\label{fig:Picture2}
\end{figure}
The mutual information shared between random variables X (RF) and Y (output) can be written as \cite{kay2011coherent}:

\begin{equation}
I(X;Y) = H(X) - H(X|Y)
\end{equation}

Where, H(X) is the Shannon entropy associated with the distribution of X and $H(X|Y)$ is the Shannon entropy associated with the conditional distribution of X given Y. It is defined as the information contained in X but not in Y \cite{kay2011coherent}. It is assumed that the mutual information is always non-negative when random variables are stochastically independent \cite{kay2011coherent}. Since we are dealing with four random variables, the conditional mutual information can be written as:

\begin{equation}
I(X; Y|Z, U) = H(Y|Z, U) - H(Y|X, Z, U)
\end{equation} 

This is the conditional mutual information shared between X and Y, having observed Z and U. It is defined as the information shared between X and Y but not shared with Z and U.\\

The four-way mutual information shared among four random variables X, Y, Z, and U can be defined as:

\begin{multline}
I(X; Y; Z; U) = I(X; Y) - I(X; Y|Z, U) = I(X; Z) - I(X; Z|Y, U) = \\ 
I(X; U) - I(X; U|Y, Z) = I(Y; Z) - I(Y; Z|X, U) = I(Y; U) - I(Y; U|X, Z)
\end{multline}

If the four-way mutual information is positive, Shannon entropy associated with the distribution of Y can be defined as \cite{kay2011coherent}:

\begin{multline}
H(Y) = I(Y; X; Z; U) + I(Y; X|Z, U) + I(Y; Z|X, U) + I(Y; U|X, Z) + H(Y; X|Z, U)
\end{multline}

In case the random variables are discrete, the integrals are replaced by summations, and the probability mass function can be written as \cite{kay2011coherent}:

\begin{equation}
H(Y) = -\int p(y)log{p(y)}dy
\end{equation}

\begin{equation}
H(Y|X) = -\int \int p(y|x)log{p(y|x)}p(x)dydx
\end{equation}

\begin{equation}
H(Y|X, Z) = -\int \int \int p(y|x, z)log{p(y|x,z)}p(x,z)dydxdz
\end{equation}

\begin{equation}
H(Y|X, Z, U) = -\int \int \int \int p(y|x, z, u)log{p(y|x,z,u)}p(x,z,u)dydxdzdu
\end{equation}

The objective function to be maximized can be defined as:

\begin{multline}
F = \phi_0 I(Y; X; Z; U) + \phi_1 I(Y; X|Z, U) + \phi_2 I(Y; Z|X, U) + \phi_3 I(Y; U|X, Z) 
+ \phi_4 H(Y; X|Z, U)
\end{multline}

$I(Y; X|Z, U)$ is the information that the output shares with the RF (X) and is not contained in the LCF and UCF units. $ I(Y; Z|X, U)$ is the information that the output shares with the LCF and not contained in the RF and UCF units. $I(Y; U|X, Z)$ is the information that the output shares with the UCF and not contained in the RF and LCF units.\\

The values of $\phi's$ are tunable within the range [-1, 1]. Different $\phi$ values allow investigating specific mutual/shared information, such that: 

\[
f(x)= 
\begin{cases}
F = I(Y; X),& \text{if } \phi_1=1, \phi_2= \phi_3=\phi_4=0\\
F = I(Y; Z),& \text{if } \phi_2=1, \phi_1= \phi_3=\phi_4=0\\
F = I(Y; U),& \text{if } \phi_3=1, \phi_1= \phi_2=\phi_4=0\\
I(Y; X; Z; U),              & \text{otherwise}
\end{cases}
\]

\begin{figure} 
	\centering
	\includegraphics[trim=0cm 0cm 0cm 0cm, clip=true, width=0.22\textwidth]{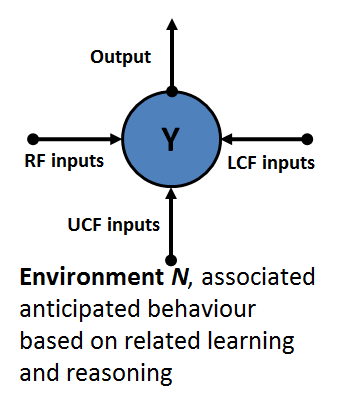}
	\caption{General human behavioural modelling in any environment \textit{N} using the conscious neural  structure.}
	\label{fig:Picture2}
\end{figure}
\begin{figure} 
	\centering
	\includegraphics[trim=0cm 0cm 0cm 0cm, clip=true, width=0.72\textwidth]{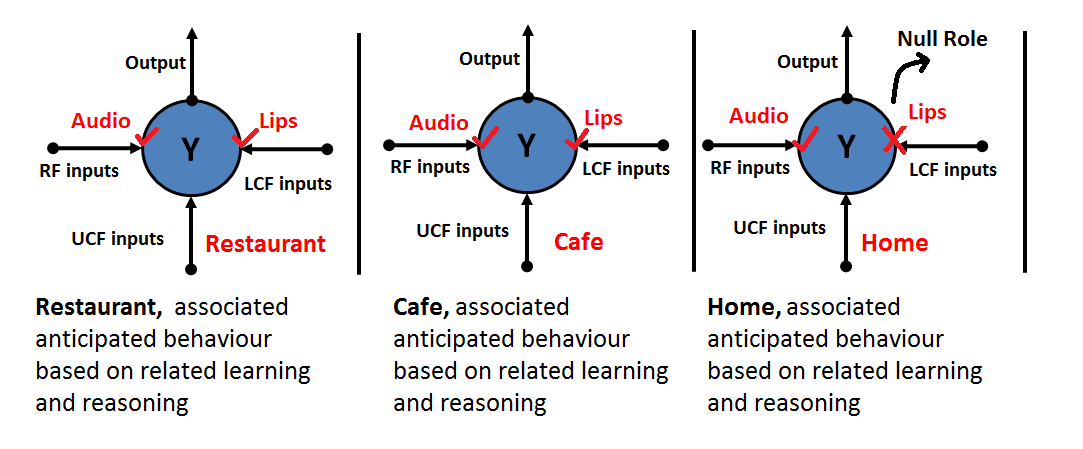}
	\caption{Human AV speech processing model in three different environments using  the conscious neural structure. Please note that in the first two environments, LCF (visual information from lip movements) has a modulatory role, but in the third environment it has a Null role. }
	\label{fig:Picture2}
\end{figure} 
\section{Human behavioural modelling}
Contextual identification and transition are two difficult problems. Given any desired human behaviour to be modelled, a set of appropriate contexts could be identified and grouped together to develop a computationally efficient model (given a broader understanding of the task in hand). According to the proposed theory in this paper, a combination of RF, LCF, and UCF can help modelling different human behaviours more precisely. A general human  behavioural model is depicted in Figure 6. The two distinctive input variables RF and LCF are defining the incoming sensory inputs (e.g., vision and sound), whereas the UCF input is defining the specific situation (outside world) and associated anticipated behaviour. In the proposed neural structure, the role of RF and LCF changes with respect to the outside environment (UCF). To further illustrate the proposed theory, following are the two case studies.


\subsection{Case Study 1: Human AV speech processing}
Human performance for speech recognition in a noisy environment is known to be dependent upon both aural and visual cues, which are combined by sophisticated multi-level integration strategies to improve intelligibility \cite{adeel2018Lip}. The multimodal nature of the speech is well established  in the literature, and it is well understood how speech is produced by the vibration of vocal folds and configuration of the articulatory organs. The correlation between the visible properties of the articulatory organs (e.g., lips, teeth, tongue) and speech reception has been previously shown in numerous behavioural studies \cite{sumby1954visual}\cite{summerfield1979use}\cite{mcgurk1976hearing}\cite{patterson2003two}. Therefore, a clear visibility of some articulatory organs could be effectively utilized to extract a clean speech signal out of a noisy audio background. Figure 7 depicts the audio-visual speech processing in three different surrounding environments: Restaurant, Cafe, and Home. In any of the environments, multisensory information (audio and visual cues) is available all the time but their optimal utilization depends on the outside environment. For example, in a busy cafe and restaurant environments (multi-talker speech perception), if there is a high background noise, our brain automatically utilizes other modalities (such as lips, body language, and facial expressions) to perceive speech or the conveyed message. Therefore, based on the information provided by the UCF (i.e. outside environment), the roles of LCF and RF are defined. For example, both RF and LCF are active in the first two scenarios (i.e. LCF modulates RF), whereas in the Home scenario (with little or zero noise), lip-reading is less effective for speech enhancement and indeed of no importance (Null role). This phenomenon is shown in our previous work at the network level \cite{adeel2018contextual}, where we showed that lip-reading driven speech enhancement significantly outperforms benchmark audio-only (A-only) speech enhancement approaches (such as spectral subtraction and log-minimum mean square error) at low signal-to-noise ratios (SNRs). However, at low levels of background noise, visual cues become less effective. 


\begin{figure} [htb!]
	\centering
	\includegraphics[trim=0cm 0cm 0cm 0cm, clip=true, width=1\textwidth]{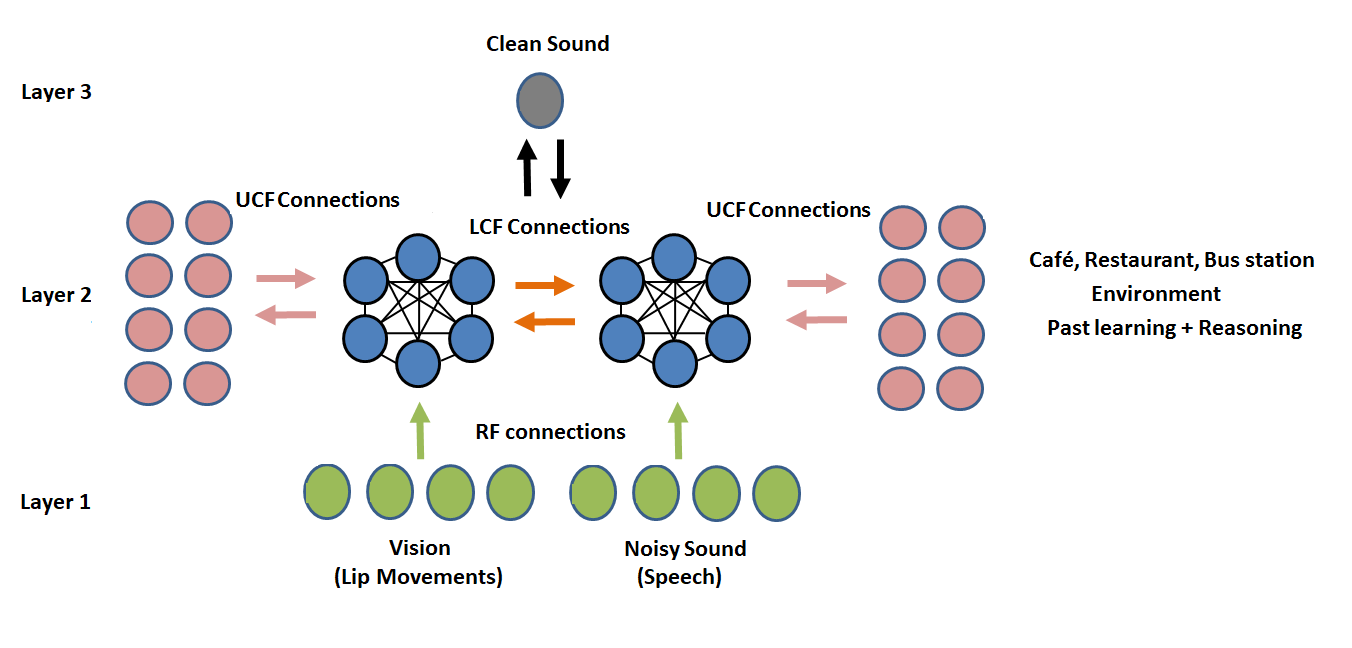}
	\caption{Audio-visual speech processing: Shallow SCNN - Three streams-three layered multisensory multiunit network of several similar conscious neurons (where the unit in one stream is connected to all other units in the neighbouring streams). Different surrounding environments defining the respective anticipated behaviour and establishing the roles of incoming multisensory signals.}
	\label{fig:Picture2}
\end{figure} 
\begin{figure}  [htb!]
	\centering
	\includegraphics[trim=0cm 0cm 0cm 0cm, clip=true, width=0.85\textwidth]{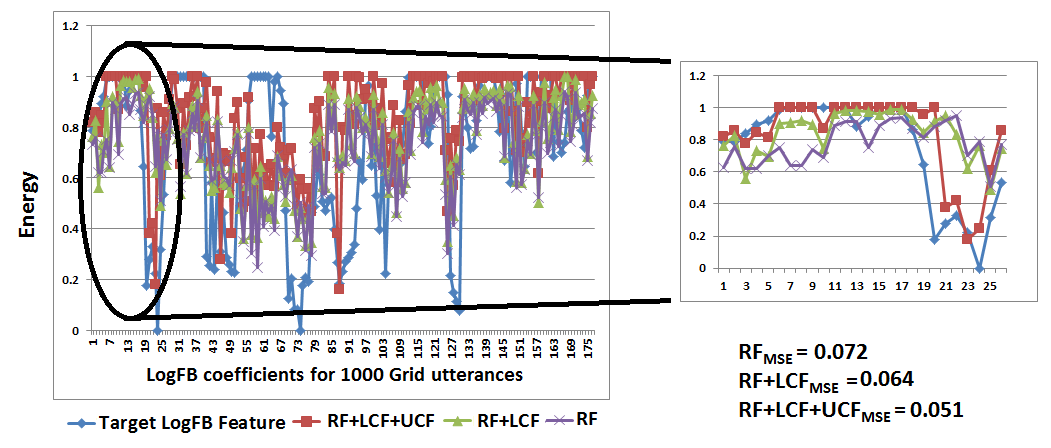}
	\caption{Neural level shallow SCNN performance: Results for A-only (RF), AV (RF+LCF), and AV with UCF, considering 3 prior AV coefficients. It is to be noted that RF+LCF+UCF model outperforms both RF-only and RF+LCF-only models. The data samples include 2D-logFB (speech) and 2D-DCT  (lip movements) coefficients for 1000 utterances from Grid and ChiME3 Corporas (Speaker 1 of the Grid). The number of clean logFB audio features are 22$\times$205,712. The combined noisy logFB audio features for -12dB, -9dB, -6dB, -3dB, 0dB, 3dB, 6dB, 9dB, and 12dB SNRs are 22$\times$205,712. Similarly, the DCT visual features are 25$\times$205,712 in total. For UCF modelling, five dynamic real-world commercially-motivated scenarios are considered: cafe, restaurant, public transport, pedestrian area, and home. Please note that a particular SNR range defines a particular environment (UCF), represented by a unique pattern using one-hot-encoding.}
	\label{fig:Picture2}
\end{figure} 
\begin{figure}[!htb]  
	\centering
	\includegraphics[trim=0cm 0cm 0cm 0cm, clip=true, width=0.80\textwidth]{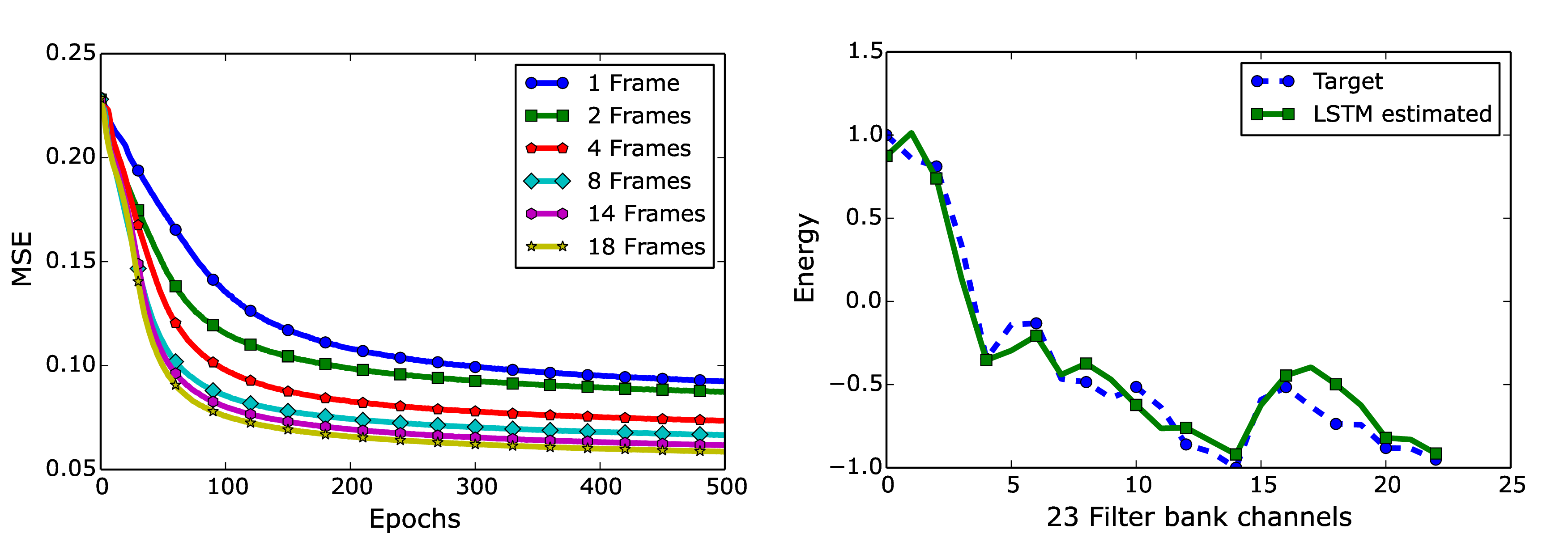}
	\caption{Network level lip-reading driven deep learning performance: Stacked LSTM validation results for different prior visual frames. The figure on left presents an overall behaviour of an LSTM model when different number of previous visual frames are added. The figure on right presents estimated clean logFB audio features using 14 prior visual frames (i.e. actual normalized energy in each of the 23 frequency bands (target, estimated)). The LSTM network had total 550 LSTM cells in the hidden layer with 23 output neurons \cite{adeel2018Lip}.}
	\label{fig:Picture2}
\end{figure} 

To test our proposed theory and SCNN, three distinctive multimodal streams (lip movements as LCF, noisy speech as RF, and the outside environment as UCF) are used. We studied how LCF and UCF are helping to improve the noisy speech filtering in different noisy conditions (ranging from a noisy (-12dB SNR) to a little noisy environment (12dB SNR)). The implemented three streams-three layered SCNN is shown in Figure 8. To train the shallow SCNN, the deep problem was transformed into a shallow problem. Specifically, in our previous AV deep learning implementation \cite{adeel2018Lip}, the output layer of the LSTM network had 23 log filter-bank (FB) coefficients (i.e. frame by frame prediction). In contrast, the evaluated shallow SCNN model predicted one coefficient at a time (i.e. coefficient by coefficient prediction). In the experiments, neurons interact by exchanging the excitatory and inhibitory spiking signals probabilistically and fire when excited as explained in Section 3. For training, the benchmark AV ChiME3 corpus is used which is developed by mixing the clean Grid videos \cite{cooke2006audio} with the ChiME3 noises \cite{barker2015third} for SNRs ranging from -12dB to 12dB \cite{adeel2018contextual}. The preprocessing includes sentence alignment and incorporation of prior audio and visual frames. Prior multiple visual frames are used to incorporate temporal information. The audio and visual features were extracted using log-FB and 2-dimensional discrete cosine transform (2D-DCT) methods. More corpus related and preprocessing details are comprehensively presented in \cite{adeel2018contextual}\cite{adeel2018Lip}. Figure 9 depicts the prediction of clean logFB coefficients, where it can be seen that multimodal RF+LCF+UCF model outperformed multimodal RF+LCF (audio-visuals) and unimodal RF (audio-only) models, achieving MSE of 0.051, 0.064, and 0.072 respectively. The performance of a network level lip-reading driven deep learning approach for speech enhancement is presented in Figure 10 \cite{adeel2018Lip}. It is to be noted that the shallow SCNN with only 29 spiking conscious neurons performed comparably to deep LSTM network which had 550 hidden cell. The ongoing work includes exploiting optimized deep learning driven AV features from \cite{adeel2018Lip}\cite{adeel2018contextual} to train SCNN.


It is believed that the enhanced learning in an SCNN is due to the shared connections and shared local and universal contextual information. The SCNN discovered and exploited the associative relations between the features extracted within each of the RF, LCF, and UCF streams. We believe that the UCF helped establishing the outside environment and anticipated behaviour and defined the roles of incoming multisensory information with respect to different situations as shown in Figure 7 (e.g. the use of audio-visual cues and audio-only cues in extreme noisy conditions and relatively clean conditions respectively).  However, to further strengthen our proposed theory, we intend to study the properties of the conscious neuron(s) using advances in information decomposition methods. Specifically, we intend to quantify the suppression and attenuation using partial information decomposition methods and explore the properties of conscious neuron and its functioning in terms of four basic arithmetic operators and their various forms \cite{kay2018contrasting}. Furthermore, we aim to critically analyze how the information is decomposed into components unique to each other having multiway mutual/shared information in recurrent SCNN.  

\begin{figure} [!htb]
\centering
	\includegraphics[trim=0cm 0cm 0cm 0cm, clip=true, width=0.65\textwidth]{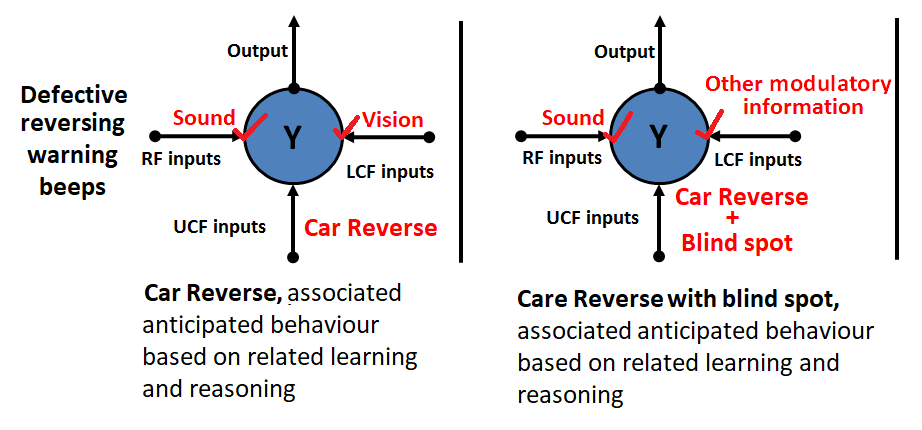}
	\caption{Driver behavioural model in two different environments using conscious neural structure. Please note that the audio cues (defective audible reversing warning beeps) are represented by ambiguous RF, coming from a defective parking sensors (partially working) which sometimes miscalculate the nearby object distance. Different surrounding environments (UCF) defining the respective anticipated behaviours and establishing the roles of incoming multisensory signals. For example, in a car reverse situation with no blind spot, the driver can't rely only on parking sensors for precise maneuvering decisions, as the audio input is ambiguous due to defective parking sensors. Therefore, in this situation, the conscious neuron defines the role of visual cues (LCF) as modulatory. In contrast, when there is a blind spot, the driver may have to rely on other modulatory signals to make an optimal decision along with the ambiguous RF.}
	\label{fig:Picture2}
\end{figure}
\begin{figure} [htb!]
	\centering
	\includegraphics[trim=0cm 0cm 0cm 0cm, clip=true, width=1\textwidth]{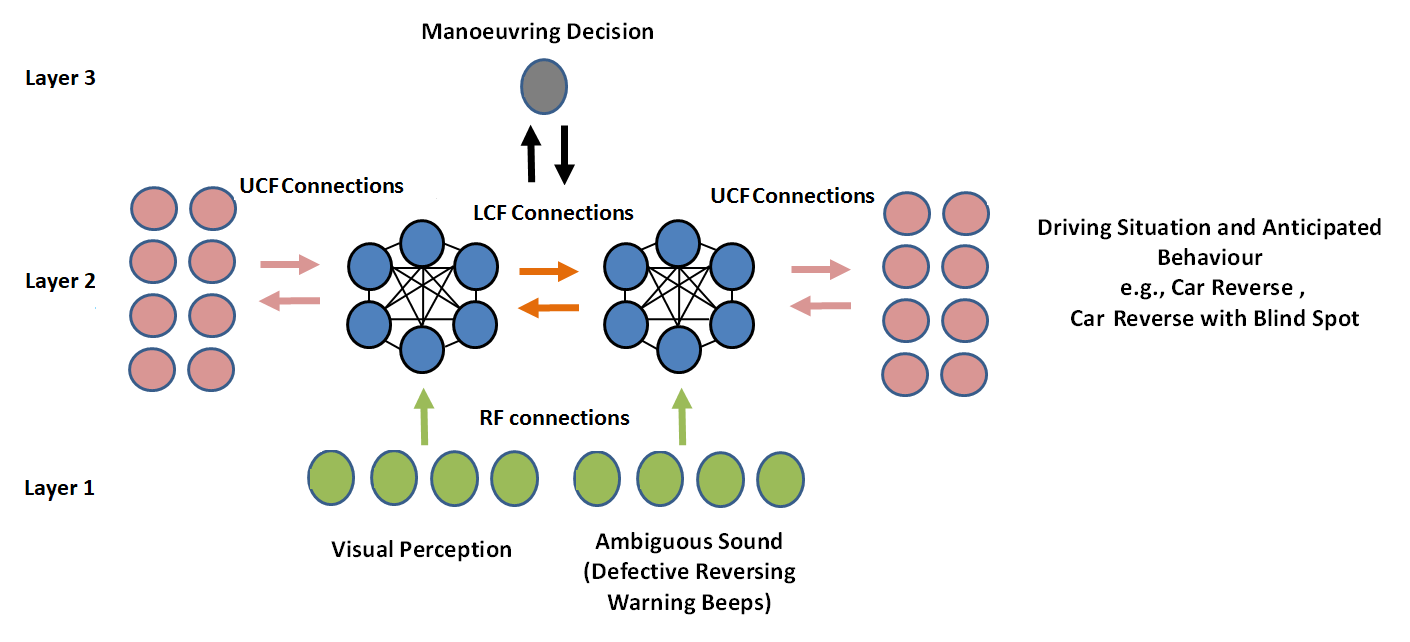}
	\caption{Driver behavioural modelling with a Shallow SCNN.}
	\label{fig:Picture2}
\end{figure} 
\subsection{Case Study 2: Driver behavioral model}
The gap between humans and machine is shrinking and scientists are trying to develop more human-like computing devices. It is becoming increasingly important to develop computer systems that incorporate or enhance situation awareness. However, methods to reduce margins of uncertainty and minimize miscommunication need further exploration. The proposed conscious neural structure and its property of originating a controlled neural command based on a precise mutisensory signals integration with respect to the external environment can help addressing these modelling challenges. For example, the proposed SCNN can help modelling a more precise driving behaviour. Figure 11 and Figure 12 depict the driving behavioural model at a neural and network level in two different surrounding environments:  car reverse with no blind spot and care reverse with blind spot. It is assumed that the parking sensors are not fully functional and at times they miscalculate the nearby object distance. Therefore, the audio input is ambiguous and the driver can't rely only on parking sensors for precise maneuvering decisions. In the first situation, where there is no blind spot, the driver leverages the complementary strengths of both AV cues. The visual cues are modulating the ambiguous audio signal (RF). In contrast, when there is a blind spot, the driver may have to rely on other modulatory signals to make an optimal decision along with the ambiguous  RF. In a nutshell, in any of the surrounding environments (UCF), multisensory information is available, but depending on the situation, the roles of incoming multisensory signals are defined to originate a precise control command (driver's maneuvering decision) complying with the anticipated behaviour. 

\section{Discussion, Research Impact, and Future Directions}
In this research, we introduced a novel theory on the role of awareness and universal context in an SCNN. The proposed theory throws light on selective amplification and attenuation of incoming multisensory information in the brain with respect to the external environment and anticipated behaviour. Specifically, it defines a guidance framework to study and model human behaviours and their underlying neural functioning in different conditions. The proposed SCNN is used to model human AV speech processing and driving behaviour. For AV speech modelling, the SCNN outperformed state-of-the-art multimodal (RF+LCF) and unimodal (RF-only) processing models. Similarly, in driver behavioural modelling, it is shown that the conscious neuron allows modelling a more precise human driving behaviour. We hypothesize that the integration of RF, LCF, and UCF helped SCNN to discover and exploit the associative relations between the features extracted within each of the RF, LCF, and UCF streams. This integration and the shared local and universal contextual information enabled enhanced learning. We believe that the inherent SCNN properties ideally place it as a powerful tool for precise behavioural modelling. However, an in-depth analysis is required to further study the properties of conscious neuron(s). In the future, we intend to use the advances in information decomposition to quantify the suppression and attenuation using partial information decomposition methods. Ongoing work also includes the development of hierarchical deep SCNN (HD-SCNN) by integrating multiple SCNNs responsible for a specific human behaviour such as audio processing, visual processing etc. For the training of HD-SCNN, we are using a theory of hypnosis for the selective training of a subnetwork (single SCNN) without affecting other already well-trained models. The testing of the proposed theory using biomedical and clinical experimental methods is also a part of our ongoing work. In subsequent sections, we present the application of SCNN in developing more human-like computing devices, low-power neuromorphic chips, and modelling sentiment and financial behaviours. 
\subsection{Research Impact}
\subsubsection{Understanding Neurodegenerative Processes using Biomedical and Clinical Methods}
Sensory impairments have an enormous impact on our lives and are closely linked to cognitive functioning. Neurodegenerative processes in Alzheimer's disease (AD) and Parkinson's disease (PD) affect the structure and functioning of neurons, resulting in altered neuronal activity \cite{liebscher2016selective}. However, the cellular and neuronal circuit mechanisms underlying this disruption are elusive. The patients with AD suffer from sensory impairment and they lack the ability to channelise awareness. Therefore, it is important to understand how multisensory integration process changes in AD and why AD patients fail to guide their actions. 

Our ongoing work includes designing an appropriate subjective testing protocol using biomedical and clinical methods to observe the role of RF, LCF, and UCF in processing and learning. For example, study AD and normal mice to observe differences in their multisensory integration processes with respect to different environmental conditions (e.g. circadian rhythms). The circadian context could be used as a UCF e.g., to define day and night along with the associated expected behaviours. The incoming multisensory information such as information from retinal ganglion cells (RGCs) could be used as RF/LCF in the proposed computational model. We believe that an integration of the proposed computational model with biomedical and clinical experiments can help understanding the underlying disrupted neural processing in different medical conditions. Specifically, it can help understanding the precise and imprecise neural firing in normal and neurodegenerative disorder patients respectively, and how different medical conditions affect the functioning of neurons. The experimental observations in the light of the proposed theory can be quantified to develop improved normal/AD/PD models. 

\subsection{Low-power Neuromorphic Chips and Internet of Things (IoT) Sensors}
The controlled firing property of the proposed conscious neural structure can help developing highly energy-efficient (low-power) neuromorphic chips and IoT sensors. The proposed SCNN inherently leverages the complementary strengths of incoming multisensory signals  with respect to the outside environment and anticipated behaviour. For example, as explained in case study 1 that in a high background noise, the conscious neuron leverages the complementary strengths of both visual and audio cues to perceive ambiguous speech. In contrast, in a low background noise, the audio cues are good enough to solve the problem. Consequently, the synaptic weights associated with the input audio cues in case of a low background noise possess high synaptic strength that leads to firing of relevant neurons and dysfunctioning of neurons associated with visual cues. This precise neural firing behaviour stops the unnecessary successive neural processing and power consumption which could be very useful in developing low-power wireless sensors. Ongoing work also includes the development of low-power neuromorphic chips and IoT sensors based on our proposed theory. 


\subsection{Other Real-World Applications}
The proposed theory can also be applied to address problems such as developing accurate financial market models, sentiment analysis models etc. The authors in \cite{kraus2017decision} proposed a decision support from financial disclosures using deep neural networks and transfer learning. Specifically, the authors used deep recurrent neural network (LSTM) to automatically extract features from ordered sequences of words in order to
capture highly non-linear relationships and context-dependent meanings. The authors demonstrated a higher directional accuracy as compared to traditional machine learning methods when predicting stock price movements in response to financial disclosures. Similar network level multimodal integration has widely been used for applications such as sentiment analysis, emotion recognition, and deception detection \cite{zou2018microblog}  \cite{gogate2017novel}\cite{gogate2017deep}. For example, the authors in \cite{zou2018microblog}  addressed the problem of ambiguous and context-aware tweets utilizing connected adjacent information with world-level and tweet level context attention. However, such implementations exploit the temporal contextual information or LCF at the network level with no integration of the overall knowledge of the problem (awareness) at the neural level, restricting accurate modelling or precise behavioral representation.

\section*{Acknowledgment}
This work was supported by the UK Engineering and Physical Sciences Research Council (EPSRC) Grant No.  EP/M026981/1 and deepCI.org. The author would like to greatly acknowledge Prof. Amir Hussain and Mandar Gogate from the University of Stirling for their contributions in implementing lip-reading driven deep learning approach and contextual AV switching for speech enhancement, which are published previously and cited here for reference. The author would also like to acknowledge Prof. Bruce Graham, Prof. Leslie Smith, Prof. Peter Hancock, and Prof. Bill Phillips from the University of Stirling, Areej Riaz from the London Business School, and Dr Mino Belle and Prof. Andrew Randall from the University of Exeter for their help and support in several different ways including appreciation, motivation, and encouragement.

\bibliographystyle{IEEEtran}
\bibliography{referdif1.bib}
\end{document}